\pdfoutput=1
\documentclass[11pt]{article}
\usepackage{nodalida2023}
\usepackage{times}
\usepackage{url}
\usepackage{latexsym}

\aclfinalcopy 
\def\aclpaperid{076} 

\title{A Survey of Corpora for\\Germanic Low-Resource Languages and Dialects}

\author{Verena Blaschke\\
\And Hinrich Schütze\\
  Center for Information and Language Processing (CIS), LMU Munich, Germany\\
  Munich Center for Machine Learning (MCML), Munich, Germany\\
  {\tt blaschke@cis.lmu.de \hspace{0.5em} inquiries@cislmu.org \hspace{0.5em} bplank@cis.lmu.de}
\And Barbara Plank\\
  }

\date{}

\usepackage{hyperref}
\usepackage{graphicx}

\usepackage[T5,T1]{fontenc} 

\usepackage{tipa}

\usepackage{microtype}

\usepackage{booktabs}
\usepackage{multirow}
\usepackage{adjustbox} 
\usepackage{siunitx}

\newcommand{\minivskip}{\vspace{0.5em}}
\newcommand{\minicolspace}{\hspace{1mm}}

\newenvironment{annotatedtable}
{%
\begin{adjustbox}{max width=\textwidth, center}
\begin{tabular}{llll@{}r@{}}\toprule
Corpus & Langs & Annotation & Size & Rep.~ \\\midrule
}
{\bottomrule
\end{tabular} 
\end{adjustbox}
}
\newcommand{\annotatedcorpus}[6]{%
#1 & #3 & #4 & #5 & #6 \\
\multicolumn{5}{l}{#2}\minivskip\\}

\newenvironment{othertable}
{
\begin{adjustbox}{max width=\textwidth, center}
\begin{tabular}{lll@{}r@{}}\toprule
Corpus & Langs & Size & Rep.~ \\\midrule
}
{\bottomrule
\end{tabular} 
\end{adjustbox}
}
\newcommand{\othercorpus}[5]{%
#1 & #3 & #4 & #5\\
\multicolumn{4}{l}{#2} \minivskip\\
}
\newcommand{\othercorpusone}[4]{%
#1 & #2 & #3 & #4 \vspace{0.5em}\\
}

\newcommand*{\pinMorphoSemParUncur}[5]{#1&#2&#3&#4&#5&}
\newcommand*{\formats}[5]{#1&#2&#3&#4&#5}

\newcommand{\https}[1]{{\href{https://#1}{\footnotesize\texttt{#1}}}}
\newcommand{\http}[1]{{\href{http://#1}{\footnotesize\texttt{#1}}}}

\newcommand{\asterisk}{*}

\usepackage{fontawesome}

\newcommand*{\doculectOrtho}{\faPencil}
\newcommand*{\phonetic}{\faPencilSquareO}
\newcommand*{\normalized}{\faParagraph}
\newcommand*{\ownOrtho}{\faFont}
\newcommand*{\audio}{\faMicrophone}

\newcommand*{\lock}{\faLock}

\newcommand*{\yes}{\faCheck}
\newcommand*{\no}{}
\newcommand*{\pin}{\faMapMarker}

\begin{document}
\maketitle

\begin{abstract}
Despite much progress in recent years, the vast majority of work in natural language processing (NLP) is on standard languages with many speakers.
In this work, we instead focus on low-resource languages and in particular
non-standardized low-resource languages. 
Even within branches of major language families, often considered well-researched, little is known about the extent and type of available resources and what the major NLP challenges are for these language varieties. 
The first step to address this situation is a systematic survey of available corpora (most importantly, annotated corpora, which are particularly valuable for NLP research).
Focusing on Germanic low-resource language varieties, we provide such a survey in this paper.  Except for geolocation (origin of speaker or document), we find that manually annotated linguistic resources are sparse and, if they exist, mostly cover morphosyntax. 
Despite this lack of resources, we observe that interest in this area is increasing: there is active development and a growing research community.
To facilitate research, we make our overview of over 80 corpora publicly available.\footnote{We share a companion website of this overview at \https{github.com/mainlp/germanic-lrl-corpora}.}
\end{abstract}

\section{Introduction}

The majority of current NLP today focuses on standard languages. Much work has been put forward in broadening the scope of NLP~\cite{joshi-etal-2020-state}, with long-term efforts pushing  boundaries for language inclusion, for example in resource creation (e.g., Universal Dependencies \cite{zeman2022ud-2-11}) and 
cross-lingual transfer research \cite{de-vries-etal-2022-make}.  
However, even within major branches of language families or even single countries, plenty of language varieties are under-researched.  

\begin{figure}[t]
    \centering
    \includegraphics[trim={17mm 5mm 13mm 30mm},clip,width=\columnwidth]{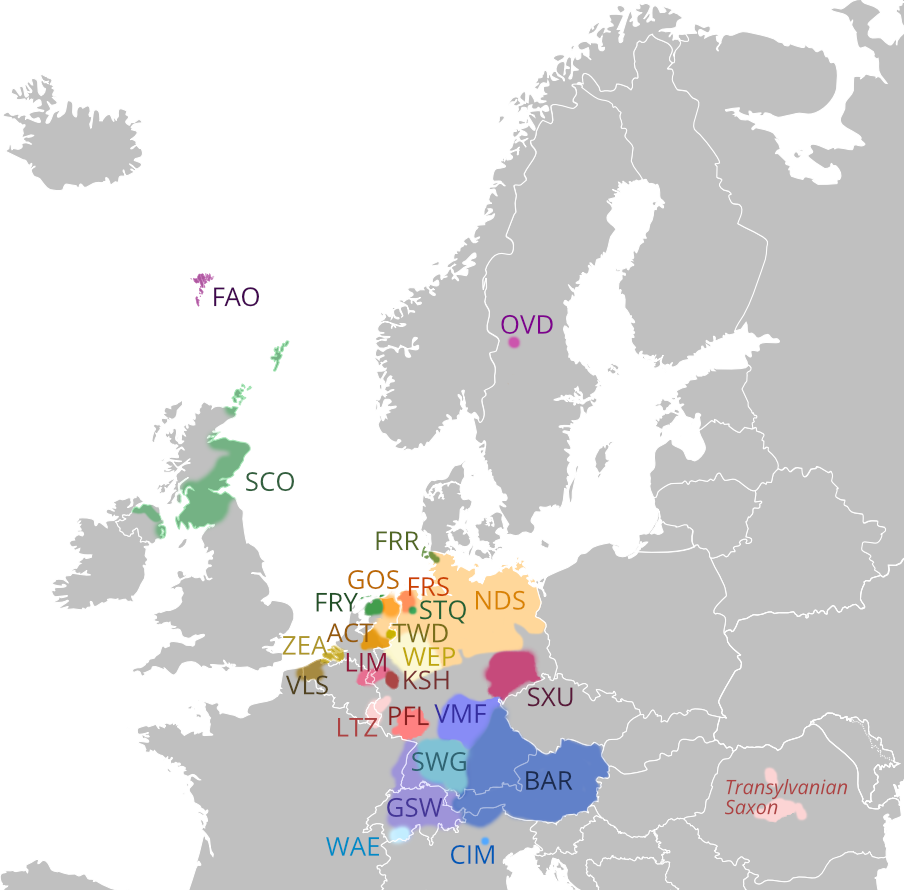}
    \caption{\textbf{Approximate locations of most of the languages discussed in this article}
    (not pictured: \textsc{pdc, yid, nor, swe, dan, eng, deu}).
    Based on a map of Europe by Marian Sigler, \href{https://creativecommons.org/licenses/by-sa/3.0/deed.en}{CC~BY-SA~3.0}.}
    \label{fig:map}
\end{figure}

Current technology lacks methods to handle scarce data and the rich variability that comes with low-resource and non-standard languages.
Nevertheless, interest in these under-resourced language varieties is growing. It is a topic of interest not only for (quantitative) dialectologists \cite{wieling2015dialectometry, nerbonne2021dialectology-cl}, but also NLP researchers, as evidenced by 
specialized 
workshops like VarDial\footnote{\https{sites.google.com/view/vardial-2023}}, special interest groups for endangered\footnote{SIGEL, \https{acl-sigel.github.io}} and under-resourced languages,\footnote{SIGUL, \http{www.elra.info/en/sig/sigul}}
and recent research on local languages spoken in Italy \cite{ramponi2022nlp}, Indonesia \cite{aji-etal-2022-one} and Australia \cite{bird-2020-decolonising}, to name but a few.

In this paper, we provide an overview of the current state of NLP corpora for Germanic low-resource languages (LRLs) and dialects, with a particular focus on non-standard variants and four dimensions: annotation type, curation profile, resource size, and (written) data representation.
We find that the amount and type of data varies by language, with manual annotations other than for morphosyntactic properties or the speaker's dialect or origin being especially rare. 
With this survey, we aim to support development of language technologies  and  quantitative dialectological analyses of Germanic low-resource languages and dialects, by making our results publicly available. 
Finally, based on the experiences we made while compiling this survey, we share recommendations for researchers releasing or using such datasets.

\section{Related surveys}

\citet{zampieri2020survey} provide an overview on research on NLP for closely related language varieties and mention a few data sets.
Recently, several surveys focusing on NLP tools and corpus linguistics data for regional languages and dialects have been released: 
for local languages in Italy \cite{ramponi2022nlp} and France \cite{leixa2014inventaire}, 
indigenous languages of the Americas \cite{mager-etal-2018-challenges},
Arabic dialects \cite{shoufan2015dialectal-arabic-nlp-survey,younes2020maghrebi,guellil2021arabic}, 
creole languages \cite{lent2022creole},
Irish English \cite{vaughan2016sociolinguistic},
and
spoken varieties of Slavic languages \cite{dobrushina2022spoken}.
Furthermore, \citet{bahr-lamberti2016ressourcen} and \citet{fischer2019regio-online}\footnote{\href{https://regionalsprache.de/regionalsprachenforschung-online.aspx}{\texttt{regionalsprache.de/\allowbreak{}regional\allowbreak{}sprachen\allowbreak{}forschung-\allowbreak{}online.\allowbreak{}aspx}}} survey digital resources for studying varieties closely related to German, although these do not necessarily fit our inclusion criteria (cf. Section~\ref{sec:methodology}).

\section{Language varieties}

Our survey contains corpora for more than two dozen Germanic low-resource varieties, selected based on dataset availability (Appendix~\ref{sec:by-language} contains the full list).
We focus on specialized corpora showcasing regional variation, but not necessarily global variation.
This overview does not include any corpora for Germanic-based creoles like Naija, as those are included in the recent survey by \citet{lent2022creole}.
Figure~\ref{fig:map} shows where most of the doculects included in this survey are spoken.

\begin{table*}
\begin{adjustbox}{max width=\textwidth, center}
\begin{tabular}{l@{\hspace{4pt}}lll@{}r}\toprule
Corpus & Langs & Annotation & Size & Rep.\\\midrule

{UD Faroese OFT \cite{tyers2018faroese}}& {FAO} & {{POS (UPOS, Giellatekno-FAO),}} & {1.2k sents} & {\ownOrtho}\\
\multicolumn{2}{l}{\https{github.com/UniversalDependencies/UD\_Faroese-OFT}} & \multicolumn{2}{l}{dep (UD), morpho (UD), lemmas}\minivskip\\

{FarPaHC / UD Faroese FarPaHC} & {FAO} & {POS (mod.\ Penn-h, UPOS),} & {53k (FarP.) /} & {\ownOrtho}\\
\multicolumn{2}{l}{\cite{farpahc,rognvaldsson-etal-2012-icelandic}} & {phrase struc.(mod.\ Penn-h),} & {40k (UD.) toks}\\
\multicolumn{2}{l}{\https{hdl.handle.net/20.500.12537/92}} & dep (UD), morpho (UD) \\
\multicolumn{4}{l}{\https{github.com/UniversalDependencies/UD\_Faroese-FarPaHC}} \minivskip\\

LIA Treebank / UD Norwegian NynorskLIA & {NOR \pin} & {POS (UPOS, mod.\ NDT),} & {77.7k toks (L.),} & {\normalized{} \phonetic{}\asterisk\hspace{-3pt}} \\
\cite{ovrelid-etal-2018-lia} && {dep (UD, mod.\ NDT), lemmas,} & \multicolumn{2}{l}{55k toks (UD)} \\
\multicolumn{2}{l}{\https{tekstlab.uio.no/LIA/norsk/index\_english.html}}&morpho (UD)\\
\multicolumn{4}{l}{\https{github.com/UniversalDependencies/UD\_Norwegian-NynorskLIA}}\\
\multicolumn{4}{l}{\https{github.com/textlab/spoken\_norwegian\_resources/tree/master/treebanks/Norwegian-NynorskLIA}}\minivskip\\

{NDC Treebank}
& {NOR \pin}
& {POS (mod.\ NDT),}
& {66k toks}
& {\normalized{} \phonetic{}\asterisk\hspace{-3pt}}\\
\multicolumn{2}{l}{\cite{kasen-etal-2022-norwegian,johannessen2009nordic-dialect-corpus}}  & {dep (mod.\ NDT), lemmas,}\\
\multicolumn{2}{l}{\http{tekstlab.uio.no/scandiasyn/download.html}} & morpho (mod.\ NDT)\\
\multicolumn{4}{l}{\https{github.com/textlab/spoken\_norwegian\_resources/tree/master/treebanks/Norwegian-BokmaalNDC}}\minivskip\\

\annotatedcorpus{NorDial (subset) \cite{maehlum-etal-2022-annotating}}{Contact authors \lock}
{NOR}
{POS (UPOS)}
{35+ tweets}
{\doculectOrtho}

\annotatedcorpus{POS-tagged Scots corpus }
{\cite{lameris-stymne-2021-whits} \https{github.com/Hfkml/pos-tagged-scots-corpus}}
{SCO}
{POS (UPOS)}
{1k tokens}
{\doculectOrtho/\ownOrtho}

\annotatedcorpus{TwitterAAE-UD \cite{blodgett-etal-2016-demographic}}
{\https{slanglab.cs.umass.edu/TwitterAAE}}
{ENG (AAVE)}
{Dep (UD)}
{250 tweets}
{\doculectOrtho}

{UD Frisian/Dutch Fame}
& {FRY/NLD}
& {POS (UPOS), dep (UD),}
& {400 sents}
& {\ownOrtho} \\
\multicolumn{2}{l}{\cite{braggaar-van-der-goot-2021-challenges,yilmaz-etal-2016-longitudinal}} & code-switching\\
\multicolumn{4}{l}{\https{github.com/UniversalDependencies/UD\_Frisian\_Dutch-Fame}}
\minivskip\\

UD Low Saxon LSDC \cite{siewert2021lsdc-with-ud}
& {NDS \pin} 
& {POS (UPOS), dep (UD),}
& {95 sents}
& {\doculectOrtho{} \normalized\asterisk\hspace{-3pt}} \\
\multicolumn{2}{l}{\https{github.com/UniversalDependencies/UD\_Low\_Saxon-LSDC}} &  \multicolumn{2}{l}{morpho (UD), glosses (GML), lemmas}\minivskip\\

{Stemmen uit het verleden (annotated subset)}
&{VLS \pin}
&{V2 variation}
&{1.4k sents}
&{\phonetic{}}\\
\multicolumn{4}{l}{\cite{lybaert2019flemish-v2,stemmen} \https{doi.org/10.18710/NSFN2B}}\minivskip\\

{Penn Parsed Corpus of Historical Yiddish} 
&{YID}
&\multicolumn{2}{l}{POS (Penn-h), phrase struc.\ (Penn-h) \quad ca. 200k toks}
&{\asterisk}\\
\multicolumn{4}{l}{\cite{penn-hist-ydd} \https{github.com/beatrice57/penn-parsed-corpus-of-historical-yiddish}}\\

Kontatto \cite{dalnegro2020kontatto} &  {BAR} & {POS (unknown),} & {147k toks} & {\audio{} \phonetic{}}\\
{\https{kontatti.projects.unibz.it} \lock} & (South Tyrol) & lemmas (DEU)\minivskip\\

{Annotated Corpus for the Alsatian Dialects}
& {GSW (Alsatian)}
& {POS (UPOS, mod.\ UPOS),}
& {798 sents}
& {\doculectOrtho}\\
\multicolumn{2}{l}{\cite{bernhard2018alsatian-occitan-picard, bernhard2019alsatian} \https{zenodo.org/record/2536041}} & {lemmas, glosses (FRA)}\minivskip\\

\annotatedcorpus{Bisame GSW}
{\cite{bisame, millour2018ressources} \https{hdl.handle.net/11403/bisame\_gsw/v1}}
{GSW (Alsatian)}
{POS (mod.\ UPOS)}
{382 sents}
{\doculectOrtho}

{Geparstes und grammatisch annotiertes}
&{GSW (St.\ Gallen)}
&{POS (mod.\ Penn-h),}
&{100k+ toks}
&{\audio{} \phonetic}\\
\multicolumn{2}{l}{Korpus schweizerdeutscher Spontansprachdaten} & phrase struc. (Penn-h)\\
\multicolumn{2}{l}{ \cite{schoenenberger2019geparstes} (contact authors \lock)}\minivskip\\

{NOAH's corpus \cite{hollenstein2015noah-extension}} & {GSW} & {POS (mod.\ STTS,} & {115k toks} & {\doculectOrtho} \\
\multicolumn{2}{l}{\https{noe-eva.github.io/NOAH-Corpus}} & {subset: UPOS and STTS)}\minivskip\\

{UD Swiss German UZH}
&{GSW}
&\multicolumn{2}{l}{POS (UPOS, mod.\ STTS), dep (UD)\quad 100 sents}
&{\doculectOrtho}\\
\multicolumn{4}{l}{\cite{aepli2018parsing} \https{github.com/UniversalDependencies/UD\_Swiss\_German-UZH}}\minivskip\\

{WUS\_DIALOG\_GSW (subset of } & {GSW \pin} & {POS (mod.\ STTS)} & {34.7k toks} & {\doculectOrtho{} \normalized}\\
\multicolumn{4}{l}{\textit{What's up, Switzerland?}) \cite{whatsup-switzerland, Ueberwasser2017whatsup} \https{whatsup.linguistik.uzh.ch} \lock}\minivskip\\

\bottomrule
\end{tabular}
\end{adjustbox}
\caption{%
\textbf{Morphosyntactically annotated corpora.}
Abbreviations for the annotation tag sets are explained in Section~\ref{sec:morphosyntax}, as are the orthographies of entries with an asterisk (\asterisk).
Other abbreviations and symbols:
\textit{Rep.}~=~`data representation,'
\textit{dep}~=~`syntactic dependencies,' \textit{phrase struc}~=~`phrase structure,' \textit{morpho}~=~`morphological features,' 
\textit{mod.}~=~`modified,'
\textit{AAVE}~=~`African-American Vernacular English,'
\textit{GML}~=~`Middle Low Saxon,'
\textit{NLD}~=~`Dutch,' \textit{FRA}~=~`French,'
\lock{}~=~access is not immediate,
\pin{}~=~fine-grained dialect distinctions,
\phonetic{}~=~phonetic/phonemic transcription,
\doculectOrtho{}~=~pronunciation spelling, 
\ownOrtho{}~=~LRL orthography,
\normalized{}~=~normalized orthography.
}
\label{tab:annotated}
\end{table*}
\begin{table*}
\begin{annotatedtable}

{TaPaCo (subset) \cite{scherrer-2020-tapaco}} & {{NDS, GOS}} & {paraphrases} & {{1107 sents (NDS),}} & {\doculectOrtho}\\
{\https{zenodo.org/record/3707949}} &&& {122 sents (GOS)} \minivskip\\

Wenkersätze \cite{wenker1923sprachatlas, regionalsprachede} & DEU* \pin & translations & 2210 samples$\times$40 sents&\phonetic/\doculectOrtho\\
\multicolumn{2}{l}{\https{github.com/engsterhold/wenker-storage}}& (across dialects, DEU)\minivskip\\

\annotatedcorpus{SB-CH (subset) \cite{grubenmann-etal-2018-sb}}
{\https{github.com/spinningbytes/SB-CH}}
{GSW}
{sentiment}
{2.8k sents}
{\doculectOrtho}

{SwissDial \cite{swissdial}} & {GSW \pin, WAE} & {topic, translations} & {2.5--4.6~hrs$\times$8~lects~} & {\audio{} \doculectOrtho{} \normalized}\\
\multicolumn{2}{l}{\https{projects.mtc.ethz.ch/swiss-voice-data-collection} \lock} & \multicolumn{2}{l}{(across dialects and into DEU)}\minivskip\\

{xSID/SID4LR (subset) } & {GSW, BAR} & {slot and intent detection,} & {800 sents} & {\doculectOrtho}\\
\cite{van-der-goot-etal-2021-masked, 2023-findings-vardial} & (South Tyrol) & {translations (14~langs)}\\
{\https{bitbucket.org/robvanderg/sid4lr}}\\

\end{annotatedtable}

\caption{%
\textbf{Corpora with semantic annotations or parallel sentences.}
Abbreviations and symbols:
\textit{Rep.}~=~`data representation,'
\lock{}~=~access is not immediate,
\pin{}~=~fine-grained dialect distinctions,
\audio{}~=~audio,
\doculectOrtho{}~=~pronunciation spelling,
\normalized{}~=~standard orthography.
*The Wenkersätze contain samples from various German dialects, but those are not annotated directly (only the town names are shared).
}
\label{tab:semantic}
\end{table*}

\begin{table*}[]
\begin{othertable}

\othercorpus{Føroyskur talumálsbanki \cite{jacobsen2022flertalsformer}}
{\https{clarino.uib.no/corpuscle-classic/corpus-list} \lock}
{FAO}
{599.9k toks}
{\ownOrtho}

\othercorpus{BLARK 1.0 (Background text corpus) \cite{simonsen2022faroese-blark}}
{(incl. FTS \cite{fts} and Faroese Korp \cite{korp-faroese}) \http{maltokni.fo/en/resources}}
{FAO}
{25M toks}
{\ownOrtho}

{Nordic Dialect Corpus (subset) \cite{johannessen2009nordic-dialect-corpus}} & {NOR \pin, OVD \pin} & {1.9M toks (NOR),} & {\normalized{} (NOR: \phonetic{})}\\
{\http{tekstlab.uio.no/nota/scandiasyn}} && {15.7k toks (OVD)} & {(OVD: \ownOrtho)} \minivskip\\

{LIA Norsk \cite{ovrelid-etal-2018-lia}} & NOR \pin & {3.5M toks} & {\phonetic{} \normalized}\\
{\https{tekstlab.uio.no/LIA/korpus.html}} & & & {partially \audio}\minivskip\\

{Talemålsundersøkelsen i Oslo (TAUS) \cite{taus}} & NOR & {388k toks} & {\phonetic{} \normalized}\\
{\https{tekstlab.uio.no/nota/taus/}} & {(East/West Oslo) \pin}\minivskip\\

{NorDial \cite{barnes-etal-2021-nordial} (subset)}
& {NOR}
& {348 tweets}
& {\doculectOrtho}\\
\https{github.com/jerbarnes/nordial}\minivskip\\

{American Nordic Speech Corpus (CANS) \cite{johannessen-2015-corpus}} 
&{{NOR (US/Canada) \pin,}}
&{{773k toks (NOR),}}
&{\phonetic{} \normalized}\\
{\https{tekstlab.uio.no/norskiamerika/korpus.html}} & {SWE (US) \pin} & {46k toks (SWE)}\minivskip\\

{Parallel dialectal--standard Swedish data} 
&{{SWE (Finland) \pin,}}
&{86.5k tokens}
&{\phonetic{} \normalized}\\
\multicolumn{4}{l}{\cite{normalization-finland-swedish, spara-finlandssvenska} \https{zenodo.org/record/4060296}} \minivskip\\

{Danish Gigaword (subset)}
&{{DAN (Bornholm)}}
&{ca. 400k tokens}
&{unk.}\\
\multicolumn{4}{l}{\cite{stromberg-derczynski-etal-2021-danish, kjeldsen2019bornholmsk} \https{gigaword.dk}} \minivskip\\

{Scottish Corpus of Texts \& Speech (SCOTS)}
& {SCO} 
& {{(unknown how many of}} & {mix of}\\
\multicolumn{2}{l}{(subset) \cite{anderson2007scots} \https{scottishcorpus.ac.uk}} &{4.6M toks in SCO)} & {\doculectOrtho{} \normalized}\minivskip\\

{Low Saxon Dialect Classification (LSDC)}
&{NDS, WEP, FRS,}
&{88.9k sents}
&{\doculectOrtho}\\
{\cite{siewert2020lsdc} \https{github.com/Helsinki-NLP/LSDC/}} & TWD, ACT \pin \minivskip\\

{LuxId \cite{lavergne-etal-2014-automatic} \href{http://lrec2014.lrec-conf.org/en/shared-lrs/current-list-shared-lrs}{\small\texttt{lrec2014.lrec-conf.org/en/}}}
&{{LTZ/DEU/FRA}}
&{{924 sents (most}}
&{\ownOrtho}\\
{\href{http://lrec2014.lrec-conf.org/en/shared-lrs/current-list-shared-lrs}{\small\texttt{shared-lrs/current-list-shared-lrs}}} & code-switching & {with LTZ content)}\minivskip\\

{DiDi (subset) \cite{didi}} & {BAR (South Tyrol)} & {unknown} & {\doculectOrtho}\\
\multicolumn{2}{l}{\https{hdl.handle.net/20.500.12124/7}} \minivskip\\

\othercorpus{What's up, Switzerland?}
{ \cite{whatsup-switzerland, Ueberwasser2017whatsup} \https{whatsup.linguistik.uzh.ch} \lock}
{GSW \pin}
{{507k msgs / 3.6M toks}}
{\doculectOrtho}

Swatchgroup Geschäftsbericht (subset) \cite{graen2019pacoco} & GSW & 79.6k toks & \doculectOrtho \\
\https{pub.cl.uzh.ch/wiki/public/pacoco/start}\minivskip\\

\othercorpus{Text+Berg (subset) \cite{textberg, graen2019pacoco}}
{\https{textberg.ch/site/en/corpora} \lock \quad \https{pub.cl.uzh.ch/wiki/public/pacoco/start}}
{GSW}
{{156 sents / 3.1k toks}}
{\doculectOrtho}

\othercorpus{ArchiWals / CLiMAlp \cite{angster2017diwac-archiwals,gaeta2020observer}}
{\https{climalp.org} \lock}
{WAE \pin}
{80+k tokens}
{\doculectOrtho}

\end{othertable}

\caption{%
\textbf{Other curated text corpora.}
Abbreviations and symbols:
\textit{Rep.}~=~`data representation,'
\lock{}~=~access is not immediate,
\pin{}~=~fine-grained dialect distinctions,
\phonetic{}~=~phonetic/phonemic transcription,
\doculectOrtho{}~=~pronunciation spelling, 
\ownOrtho{}~=~LRL orthography,
\normalized{}~=~normalized orthography.
}
\label{tab:curated}
\end{table*}

\begin{table*}[]
\begin{adjustbox}{max width=\textwidth, center}
\begin{tabular}{ll@{}l@{\hspace{-15pt}}r@{}}\toprule
Corpus & Langs & Size & {Rep.~} \\\midrule

\multicolumn{2}{l}{BLARK 1.0 (Transcr. recordings) \cite{simonsen2022faroese-blark} \http{maltokni.fo/en/resources} \quad FAO \pin}& 100\,h & \hspace{-20pt} \audio{} \ownOrtho{} (some \phonetic{}\kern1pt)\minivskip\\

\othercorpus{Faroese Danish Corpus Hamburg (FADAC Hamburg) (subset)}
{\cite{debess2019fadac} \https{corpora.uni-hamburg.de/hzsk/de/islandora/object/spoken-corpus:fadac-0.2.dan}}
{FAO \pin} 
{31\,h}
{\audio{} \ownOrtho}

{NB Tale -- Speech Database for Norwegian \cite{nb-tale}}
&{NOR \pin}
&{365\,$\times$\,2\,min (spon.),}
&{\audio{} \phonetic{} \normalized}\\
\multicolumn{2}{l}{\https{nb.no/sprakbanken/en/resource-catalogue/oai-nb-no-sbr-31/}} & {7.6k sents (reading)} \minivskip\\

\othercorpus{Norwegian Parliamentary Speech Corpus (NPSC)}
{\cite{solberg-ortiz-2022-norwegian} \https{nb.no/sprakbanken/en/resource-catalogue/oai-nb-no-sbr-58/}}
{NOR \pin}
{140\,h}
{\audio{} \normalized}

{Diachronic Electronic Corpus of Tyneside English (DECTE)}
&{ENG (UK: Tyneside)}
&{~72\,h / 804k toks}
&{\audio{} \normalized{}}\\
\multicolumn{3}{l}{\cite{decte} \https{research.ncl.ac.uk/decte/index.htm} \lock} &  (some \phonetic{}\kern1pt) \minivskip\\

{Intonational Variation in English (IViE)}
&{ENG}
&{36\,h}
&{\audio{} \normalized}\\
{\cite{ivie} \http{phon.ox.ac.uk/files/apps/IViE/}} & (UK, Ireland) \pin \minivskip\\

{Crowdsourced high-quality UK and Ireland English Dialect speech data set}
& {ENG}
&{31\,h}
&{\audio{} \normalized}\\
{\cite{demirsahin-etal-2020-open} \https{openslr.org/83}} & (UK, Ireland) \pin \minivskip\\

{Helsinki Corpus of British English Dialects}
&{ENG (UK) \pin}
&{1\,M toks}
&{\audio{} \normalized}\\
\multicolumn{4}{l}{\cite{helsinki-dialects} \https{varieng.helsinki.fi/CoRD/corpora/Dialects/} \lock}\minivskip\\

{Nationwide Speech Project (NSP) \cite{clopper2006nsp} \https{u.osu.edu/nspcorpus}} & ENG (US) \pin
&{60\,$\times$\,1 hr}
&{\audio{} (some \normalized{}\kern1pt)}\minivskip\\

{Corpus of Regional African American Language (CORAAL)}
&{ENG (AAVE) \pin}
&{135.6 hrs / 1.5M toks}
&{\audio{} \normalized{}}\\
{\cite{coraal} \https{oraal.uoregon.edu/coraal}} \minivskip\\

\othercorpus{Common Voice Corpus 12.0 (subset) \cite{ardila-etal-2020-common}}
{\https{commonvoice.mozilla.org/en/datasets}}
{FRY}
{150\,h}
{\audio{} \ownOrtho}

\othercorpus{Frisian AudioMining Enterprise (FAME)}
{\cite{yilmaz-etal-2016-longitudinal} \https{ru.nl/clst/tools-demos/datasets/} \lock}
{FRY (some \pin)}
{18.5\,h}
{\audio{} \ownOrtho}

\othercorpus{Recordings of Dutch-Frisian council meetings}
{\cite{bentum-etal-2022-speech} \https{frisian.eu/dutchfrisiancouncilmeetings}}
{FRY}
{26\,h / 281k toks}
{\audio{} \ownOrtho}

{Corpus Spoken Frisian \cite{corpus-spoken-frisian}} {\https{www1.fa.knaw.nl/ksf.html} \lock}
&\multicolumn{2}{l}{FRY \hfill 200\,h (65\,h transcribed)}
&{\audio{} (\ownOrtho)}\minivskip\\

\othercorpus{Sprachvariation in Norddeutschland (SiN, Hamburg collection)}{\cite{schroeder2011sin, elmentaler2015sprachvariation} \https{hdl.handle.net/11022/0000-0000-7EE3-3} \lock}
{NDS, FRS, DEU}
{300\,h}
{\audio}

\othercorpus{Regional Variants of German 1 (RVG1)}
{\cite{rvg1} \https{hdl.handle.net/11022/1009-0000-0004-3FF4-3}}
{DEU\asterisk{} \pin}
{500+\,$\times$\,1\,min}
{\audio{} \phonetic{} \normalized{}}

{Zwirner-Korpus (downloadable subset)}&
{NDS, WEP, SXU, }&
{3\,h / 24.8k toks}&
{\audio{} \normalized}\\
{\cite{zwirner, dgd} \https{dgd.ids-mannheim.de} \lock} & \multicolumn{2}{l}{VMF, BAR, GSW, DEU** \pin} \minivskip\\

\othercorpus{Texas German Sample Corpus (TGSC) \cite{texas-german-sample-corpus}}
{\https{doi.org/10.18738/T8/IOX9ZA}}
{DEU (Texas)}
{13.5\,h / 75k toks}
{\audio{} \normalized} 

{Audioatlas Siebenbürgisch-Sächsischer Dialekte}
&{DEU}
&{360\,h / 450k toks}
&{\audio{} \normalized{}}\\
{\cite{asd} \https{hdl.handle.net/11022/1009-0000-0001-27B9-3} \lock} & (Trans. Saxon)*** & & (some \phonetic{}\kern1pt) \minivskip\\

CABank Yiddish Corpus \cite{newman2015discourse} \https{ca.talkbank.org/access/Yiddish.html} & YID (New York)
&{1 hr}
&{\audio{} \phonetic{}}\minivskip\\

\othercorpusone{SXUCorpus \cite{herms-etal-2016-corpus} Contact authors \lock}
{SXU \pin}
{500\,min / 70k toks}
{\audio{} \normalized}

\othercorpusone{Kontatti \cite{ghilardi2019eliciting} \https{kontatti.projects.unibz.it} \lock}
{BAR (S.\ Tyrol), CIM}
{~unknown}
{\audio{} \normalized}

\othercorpus{ArchiMob \cite{Scherrer2019ArchiMob}}
{\https{spur.uzh.ch/en/departments/research/textgroup/ArchiMob.html} (audio files: \lock)}
{GSW \pin}
{70\,h}
{\audio{} \phonetic{} \normalized}

\othercorpusone{SDS-200 \cite{pluss-etal-2022-sds} \https{swissnlp.org/datasets/} \lock}
{GSW}
{200\,h}
{\audio{} \normalized}

\multicolumn{2}{l}{Swiss Parliaments Corpus \cite{pluss2021swiss-parliaments} \https{www.cs.technik.fhnw.ch/i4ds-datasets} \quad GSW}
&{293\,h}
&{\audio{} \normalized}\minivskip\\

\multicolumn{2}{l}{Gemeinderat Zürich Audio Corpus \cite{pluss2021swisstext} \https{www.cs.technik.fhnw.ch/i4ds-datasets} \quad GSW}
&{1208\,h}
&{\audio{}} \minivskip\\

{All Swiss German Dialects Test Set \cite{pluss2021swisstext}}
& {GSW, WAE \pin}
& {13\,h /}
& {\audio{} \normalized} \\
{\https{www.cs.technik.fhnw.ch/i4ds-datasets}} & &  5.8k utterances \minivskip\\

\multicolumn{2}{l}{Walliserdeutsch/RRO \cite{garner2014walliserdeutsch-rro,garner2014automatic} {\https{zenodo.org/record/4580286} \lock} {WAE}} &
{8.3\,h} &
{\audio{} \doculectOrtho} \minivskip\\

\bottomrule
\end{tabular} 
\end{adjustbox}

\caption{%
\textbf{Other audio corpora.}
Abbreviations and symbols:
\textit{Rep.}~=~`data representation,'
\lock{}~=~access is not immediate,
\pin{}~=~fine-grained dialect distinctions,
\audio{}~=~audio,
\phonetic{}~=~phonetic/phonemic transcription,
\doculectOrtho{}~=~pronunciation spelling, 
\ownOrtho{}~=~LRL orthography,
\normalized{}~=~normalized orthography.
\asterisk{}It is unclear whether the {RVG1} recordings are in regionally accented (Standard) German or whether they are in Low Saxon, Bavarian and other regional languages spoken in Germany, Switzerland, Austria and Northern Italy.
**The {Zwirner-Korpus} contains samples from various dialects spoken in what used to be West Germany.
***Transylvanian Saxon is a variety of Moselle Franconian that does not have its own ISO code. It is more closely related to Luxembourgish than to Standard German. 
}
\label{tab:audio}
\end{table*}

\begin{table*}[]
\begin{adjustbox}{max width=\textwidth, center}
\begin{tabular}{ll}\toprule
Corpus & Languages and sizes \\\midrule

{Tatoeba (subset; with $>100$ sents)}
& {in sentences: NDS (18.1k), YDD (12.8k), GOS (5.7k), FRR (2.9k), }\\
{\https{tatoeba.org/en/downloads}} & {SWG (1.9k), LTZ (884), FRY (641), GSW (474), FAO (417), BAR (227)} \minivskip\\

{Ubuntu \https{opus.nlpl.eu/Ubuntu.php}} & {in toks: NDS (35.3k), FRY (22.4k), FAO (20.2k), LIM (18.4k), LTZ (17.0k)}\minivskip\\

{KDE4 \https{opus.nlpl.eu/KDE4-v2.php}} & {NDS (1.1M toks), FRY (0.3M toks), LTZ (28.8k toks)}\minivskip\\

{GNOME \https{opus.nlpl.eu/GNOME.php}} & {NDS (0.7M toks), LIM (0.4M toks), FRY (55.7k toks)}\minivskip\\

\multicolumn{2}{l}{Mozilla-I10n \https{mozilla-l10n/mt-training-data} \quad FRY (0.4M toks), LTZ (6.9k toks)}\minivskip\\

\multicolumn{2}{l}{QED \cite{abdelali-etal-2014-amara} \https{opus.nlpl.eu/QED.php} \quad LTZ (19.2k toks), FAO (6.4k toks)}\minivskip\\

\multicolumn{2}{l}{TED2020 \cite{reimers-gurevych-2020-making} \https{opus.nlpl.eu/TED2020.php}  LTZ (1.7k toks)} \minivskip\\

\midrule

{Danish Gigaword (subset) } 
&{{DAN (South Jutish)} (ca. 20k tokens)}\\ 
\multicolumn{2}{l}{\cite{stromberg-derczynski-etal-2021-danish} \https{gigaword.dk}}\minivskip\\

\multicolumn{2}{l}{SwissCrawl \cite{linder2020crawler} 
\https{icosys.ch/swisscrawl} \lock \quad GSW (500k+ sents)}\minivskip\\

\multicolumn{2}{l}{SB-CH \cite{grubenmann-etal-2018-sb}
\https{github.com/spinningbytes/SB-CH} \lock \quad GSW (ca.\ 116k sents)}\minivskip\\

\multicolumn{2}{l}{SwigSpot \cite{linder2018swigspot} \https{github.com/derlin/SwigSpot\_Schwyzertuutsch-Spotting} \quad GSW (8k sents)}\minivskip\\

{Web to Corpus (W2C) (subset)} &
{in\,MB: YID (125), FAO (102), LTZ (81), FRY (72), SCO (35), }\\
\multicolumn{2}{l}{\cite{w2c, majlis-zabokrtsky-2012-language} \https{hdl.handle.net/11858/00-097C-0000-0022-6133-9} \quad NDS (24), LI (20)}\minivskip\\

\multicolumn{2}{l}{CC-100 (subset) \cite{wenzek-etal-2020-ccnet} \https{data.statmt.org/cc-100/} \quad FRY (174\,MB), YID (51\,MB), LIM (8.3\,MB)}\minivskip\\

{OSCAR (subset) \cite{abadji-etal-2022-towards}}
& {in toks: YID (14.3M), FRY (9.8M),}\\
\multicolumn{2}{l}{\https{oscar-project.github.io/documentation/} \lock \quad LTZ (2.5M), NDS (1.6M), GSW (34k)}\minivskip\\

{Wikipedia (subset) \https{dumps.wikimedia.org}}&{discussed in detail in Appendix~\ref{sec:wikistats}}\\

\bottomrule
\end{tabular} 
\end{adjustbox}

\caption{%
\textbf{Uncurated corpora.} \lock{}~=~Access not immediate.
The corpora in the top section contain parallel sentences with many translations and are (also) distributed via the OPUS project \cite{tiedemann-2012-parallel}.
}
\label{tab:uncurated}
\end{table*}

\section{Methodology}
\label{sec:methodology}

Similarly to \citet{ramponi2022nlp},
we search for corpora on several online repositories for language resources:
the CLARIN Virtual Language Observatory \cite{clarin-vlo},
the LRE Map \cite{lre-map},
the European Language Grid \cite{rehm2020elg}
OLAC \cite{olac},
ORTOLANG \cite{ortolang},
and
the Hamburg Centre for Language Corpora.%
\footnote{%
\https{vlo.clarin.eu};
\https{lremap.elra.info};
\https{live.european-language-grid.eu};
\http{www.language-archives.org};
\https{www.ortolang.fr/market/corpora};
\href{https://corpora.uni-hamburg.de/hzsk/en/repository-search}{\texttt{corpora.\allowbreak{}uni-hamburg.de/hzsk/en/\allowbreak{}repository-\allowbreak{}search}}
}
We also search for corpora on Zenodo and on Google Dataset Search, and look
for mentions of corpora in articles hosted by the ACL Anthology and on ArXiv.%
\footnote{%
\https{zenodo.org};
\href{https://datasetsearch.research.google.com}{\texttt{datasetsearch.research\allowbreak{}.google\allowbreak{}.com}};
\https{aclanthology.org};
\https{arxiv.org}
}
We search for mentions of the word ``dialect'' and the names of various Germanic low-resource languages.

We use the following inclusion criteria:
\begin{itemize}
    \item The corpus is accessible to researchers (immediately via a website, or indirectly through a request form or via contact information),\footnote{The latter case is indicated with a lock \lock{} in the tables.} and this 
    is indicated on the corpus website or in a publication accompanying the corpus.
    \item The corpus can be downloaded easily (does not require scraping a website) and does not require extensive pre-processing (we are interested in file formats like XML, TSV or TXT rather than PDF).
    \item The data are of a high quality, e.g., we ignore OCR'ed corpora that were not carefully cleaned.
    \item The corpus (mainly) contains full sentences or utterances,\footnote{This excludes word lists and some heavily preprocessed corpora, like the one by \citet{hovy-purschke-2018-capturing}, which is lemmatized and does not contain stop words.} and the data were (mainly) produced in the past century.
\end{itemize}

We base this survey only on the versions of corpora that are easily accessible to the research community; e.g., if a corpus contains audio material, but only the written material is available for download (and thus as a data source for quantitative research), the corpus is treated as a text corpus.%
\footnote{This is not a rare scenario, as the audio versions might 
contain more personally identifying information (like the voice of someone from a small speaker population), and it requires more work to censor locations or personal names in audio data than in text data \cite[see also][]{seyfeddinipur2019access}.}

\section{Corpora}

Most of the language varieties we survey have no or only a very recent written tradition.
This unique challenge is reflected in the different written formats used to represent the data (if the corpora contain any written material at all) and concerns both the transcription of audio data \cite{tagliamonte2007real-language,gaeta2022corpus} as well as the elicitation of written data \cite{millour2020newlywritten}.
We opted to discern between audio data~\audio{} and the following written variants: standard orthographies (of the doculects themselves where existing~\ownOrtho{} (e.g., West Frisian orthography), or of a closely related higher-resource language otherwise~\normalized{}), ad-hoc pronunciation spelling (by speakers of the doculect)~\doculectOrtho, and phonetic or phonemic transcriptions (by linguists)~\phonetic.
Appendix~\ref{sec:transcriptions} provides examples.

The following corpora are sorted by annotation and curation type.
For an overview sorted by language, see Appendix~\ref{sec:by-language}.
Some of the corpora share the same data sources.
Appendix~\ref{sec:overlaps} lists the cases where we are aware of such overlaps.

\subsection{Annotated corpora}

This section only includes corpora with manual (or manually corrected) annotations.

\subsubsection{Morphosyntactic annotation}
\label{sec:morphosyntax}

Table~\ref{tab:annotated} provides an overview of datasets with morphosyntactic annotations.
These mostly contain part-of-speech (POS) tags and/or syntactic dependencies.
Such annotations are, for instance, of interest to dialectologists studying morphosyntactic variation 
\cite[see for example][]{lybaert2019flemish-v2}.
Automatically generating high-quality morphosyntactic annotations for non-standard and/or low-resource data is not trivial, and the more annotated data are available for training, the better the results tend to be \cite{schulz2019gmh-pos, scherrer2019digitising-gsw}.

The annotation standards tend to either be general and cross-linguistically applicable (inviting comparisons between languages), or to be very specific to the language variety at hand.
In the former case, annotations follow the guidelines from the Universal Dependencies project \cite{zeman2022ud-2-11} (UD, UPOS).
In the latter case, tag sets created for a (usually closely related) higher-resource language are modified so that they capture the lower-resource language variety's idiosyncrasies.
These specialized tag sets are based on:
the annotations of the Giellatekno project \cite{wiechetek-etal-2022-unmasking},
the annotations developed for the Penn Parsed Corpora of Historical English \mbox{(Penn-h),}\footnote{%
\href{https://ling.upenn.edu/hist-corpora/annotation/index.html}{\texttt{ling.upenn.edu/hist-corpora/\allowbreak{}annotation\allowbreak{}/\allowbreak{}index.html}}
}
the tag set of the Norwegian Dependency Treebank (NDT) \cite{solberg-etal-2014-norwegian} (based on the Oslo-Bergen Tagger's tag set, OBT, \cite{johannessen2012obt+}),
and
the Stuttgart-Tübingen tag set (STTS) \cite{stts}.

Most of the annotated corpora are presented only in one written form, typically either written in a standard orthography or pronunciation spelling.
Some cases (marked with an asterisk\asterisk{} in the table) require further explanation:
The Norwegian LIA and NDC treebanks \cite{ovrelid-etal-2018-lia, kasen-etal-2022-norwegian} use normalized orthographies (Nynorsk and Bokmål, respectively), but aligned versions of the original phonetic and orthographic transcriptions can be downloaded from the Tekstlab links in the table.
The sentences in the UD Low Saxon LSDC treebank \cite{siewert2021lsdc-with-ud} are presented both in the original ad-hoc pronunciation spelling and in a recently proposed orthography for Low Saxon, \textit{Nysassiske Skryvwyse}.
The Yiddish corpus \cite{penn-hist-ydd} is romanized, partially according to the YIVO transliteration system, and partially in a non-systematic manner.

\subsubsection{Semantic annotation and parallel sentences}

Very few resources with other types of annotations exist; we were able to find only five (Table~\ref{tab:semantic}), all of which have very different kinds of annotations: sentiment or topic classification, intent detection and slot-filling, 
translations and paraphrases.

\subsubsection{Dialect annotation}

Many corpora contain detailed annotations on the dialect area (or more precise location) an utterance's speaker or the author of a document is from.
Such information is important for linguistic research comparing related dialects \cite{wieling2015dialectometry}, 
for comparing the results of traditional and quantitative dialectological approaches \cite[e.g.][]{heeringa2009measuring}
and
for evaluating whether NLP systems perform differently on different closely related language varieties \cite{ziems2022multivalue}.
Since corpora with such annotations belong to all of the categories of curated datasets in this survey, they are not presented on their own, but instead marked with a pin symbol \pin{} elsewhere.

\subsection{Other curated corpora}

\subsubsection{Text corpora}

Table~\ref{tab:curated} presents unannotated written corpora of low-resource languages like Elfdalian or Faroese, and corpora that showcase dialectal variation through phonetic transcriptions or pronunciation spelling.
(While variation also occurs on linguistic levels encoded in normalized text written in standard orthographies --~lexical, syntactic or pragmatic variation~-- we focus on phonological variation, as this is where specialized corpora are required.)

\subsubsection{Audio corpora}

In this survey, our focus lies on written resources, and as such, this selection of audio corpora is not exhaustive.\footnote{%
Additional corpora documenting variation in spoken English can be found via the SPADE project \cite{spade}.}
However, many of the language varieties surveyed in this article are predominately spoken rather than written.
Creating language technology for unwritten languages is a topic of interest for NLP researchers \cite{scharenborg2020unwritten}, and this is also reflected by the number of recently created speech corpora for Germanic LRLs.

Many of the audio corpora (Table~\ref{tab:audio}) fall into one of two categories: recordings created for dialectological research, and post-hoc collections of already existing audio data (like radio broadcasts or public recordings of council meetings).
Most of the audio corpora are at least partially transcribed, typically according to a standard orthography.

\subsection{Uncurated text corpora}
\label{sec:uncurated}

A final type of corpus are uncurated text collections (Table~\ref{tab:uncurated}).
This includes data coming from community-based data collection efforts unrelated to research projects (Wikipedia, Tatoeba) and open-source translations of (mostly) user interfaces, as well as web-crawled data.

It is important to note that there are quality issues with web-crawled corpora, especially for low-resource languages \cite{kreutzer-etal-2022-quality}.\footnote{%
However, see \citet{artetxe2022corpus-quality} for an argument that the linguistic quality of a corpus might not be the most important factor for all downstream applications.}
Both CC-100 \cite{wenzek-etal-2020-ccnet} and OSCAR \cite{abadji-etal-2022-towards} are cleaned versions of CommonCrawl\footnote{\https{commoncrawl.org}}\kern-0.5pt -- and \citet{abadji-etal-2022-towards} specifically remark on the low quality of the low-resource language data in that dataset.

Some of the translated corpora also have quality issues: the Low Saxon Ubuntu and GNOME corpora \cite{tiedemann-2012-parallel} both also contain some Standard German content.
We exclude subcorpora that contain mostly foreign language or non-linguistic material (for instance, the West Flemish QED subcorpus \cite{abdelali-etal-2014-amara, tiedemann-2012-parallel}).

Wikipedia has editions in many Germanic low-resource languages and at different activity and contributor levels, as we survey in  Appendix~\ref{sec:wikistats}. 
Projects extend wiki dumps with automatically inferred annotations \cite{pan-etal-2017-cross,schwenk-etal-2021-wikimatrix}, or release automatically aligned German--Alemannic/Bavarian bitext \cite{artemova-plank-2023-low}.\footnote{\href{https://github.com/mainlp/dialect-BLI}{\texttt{github.com/mainlp/dialect-BLI}}}
The linguistic quality of LRL wikis is not always very high -- the Scots Wikipedia made the news in 2020, when attention was brought to the fact that half of that wiki's articles had been created/edited by a non-Scots speaker writing in a parody of Scots \cite{guardian-scots-wiki}.
Quality issues should be taken into account when working with data from small wikis without much oversight, e.g., with data or tools based on the Scots Wikipedia before clean-up started in fall 2020.\footnote{%
E.g., Scots is included in the language list of mBERT \cite{devlin-etal-2019-bert}, which was trained on Wikipedia data in 2019: 
\href{https://github.com/google-research/bert/blob/master/multilingual.md}{\texttt{github.com/\allowbreak{}google-research/\allowbreak{}bert/\allowbreak{}blob/\allowbreak{}master/\allowbreak{}multilingual.md}}
}

\section{Outlook}

Creating NLP resources and technology for LRLs is an active field.
At the time of writing this paper,  several additional resources were concurrently under construction or revision: 
\textit{UD Frisian Frysk}, a treebank for West Frisian \cite{heeringa2021ud-frisian},\footnote{%
\href{https://github.com/UniversalDependencies/UD_Frisian-Frysk}{\texttt{github.com/UniversalDependencies/\allowbreak{}UD\_Frisian-Frysk}}
}
\textit{Boarnsterhim Corpus}, a West Frisian audio corpus \cite{sloos-etal-2018-boarnsterhim},\footnote{%
\href{https://taalmaterialen.ivdnt.org/download/tstc-boarnsterhimcorpus1-0}{\texttt{taalmaterialen.ivdnt.org/download/\allowbreak{}tstc-boarnsterhimcorpus1-0}}
}
\textit{Schweizerdeutsches Mundartkorpus}, a Swiss German text corpus \cite{weibel2020compiling},\footnote{\https{chmk.ch/de/info\_all}}
and 
the \textit{Corpus of Southern Dutch Dialects}  \cite{breitbarth2018gcnd}.\footnote{\https{gcnd.ugent.be}}
Community-based projects are also being actively developed: many of the small Wikipedias have active editors (Appendix~\ref{sec:wikistats}), as do many of the Tatoeba collections. We welcome contributions to our companion website to track such progress.

Speaker populations of LRLs are not a monolith. Accordingly, different speaker communities have different interests in terms when it comes to the development of language technologies \cite{lent2022creole}. The creation of downstream technologies made for public use should be made in accordance of the wishes and needs of the relevant speaker communities \cite[see also][]{bird-2022-local}.

We make the following \textbf{recommendations} for researchers who \textit{work} with LRL datasets:
\begin{itemize}
    \item Investigate the quality of uncurated data, as it might be especially poor for LRLs.
    \item Check whether (pre-)training, development and test data are truly from independent datasets -- the dearth of high-quality LRL data means that datasets may be likely to overlap.
    \item Consider quantitative work by dialectologists and sociolinguists who might not publish in typical NLP venues.
\end{itemize}

To researchers who \textit{create} such datasets, we recommend to:
\begin{itemize}
    \item Document the transcription principles (if the data were originally in an audio format) / if any standardized orthographies were used (if the language variety does not have an official orthography).
    \item The low number of available high-quality datasets per language variety means that the impact of losing such a resource is much greater. Therefore, please upload your corpus to an archive geared towards long-term data storage (like the CLARIN Language Resource Inventory,\footnote{%
    \href{https://clarin.eu/content/language-resource-inventory}{\texttt{clarin.eu/content/language-resource-\allowbreak{}inventory}}
    } the LRE Map or Zenodo).
    \item Provide easy-to-find documentation with details on the corpus size, data sources and the annotation procedure.
\end{itemize}

\section{Conclusion}

We have presented an analysis of over 80 corpora containing data in Germanic low-resource languages, with a focus on non-standardized or only recently standardized varieties.
We additionally share the corpus overview on a public companion website (\href{https://github.com/mainlp/germanic-lrl-corpora}{\texttt{github.com/\allowbreak{}mainlp/\allowbreak{}germanic-\allowbreak{}lrl-\allowbreak{}corpora}}) that can easily be updated as more language resources are released.

\section*{Acknowledgements}
We thank the anonymous reviewers as well as the members of the MaiNLP research lab for their constructive feedback.
This research is supported by European Research Council (ERC) Consolidator Grant DIALECT 101043235.
This work was partially funded by the ERC under the European Union's Horizon 2020 research and innovation program (grant 740516).

\bibliographystyle{acl_natbib}
\bibliography{references/general.bib,
references/north-germanic.bib,
references/low-saxon.bib,
references/english.bib,
references/frisian.bib,
references/dutch.bib,
references/ud.bib,
references/yiddish.bib,
references/swiss-german.bib,
references/upper-german.bib
}

\appendix

\section{Resources by language}
\label{sec:by-language}

We include the languages associated with the ISO~639-3 codes \textsc{fao}~(Faroese),
\textsc{ovd}~(Elfdalian),
\textsc{sco}~(Scots),
\textsc{frr}~(North Frisian),
\textsc{fry}~(West Frisian),
\textsc{stq}~(Saterland Frisian),
\textsc{nds}~(Low Saxon),
\textsc{frs}~(East Frisian Low Saxon),
\textsc{gos}~(Gronings),
\textsc{twd}~(Twents),
\textsc{act}~(Achterhoeks),
\textsc{wep}~(Westphalian),
\textsc{zea}~(Zeelandic),
\textsc{vls}~(West Flemish),
\textsc{ltz}~(Luxembourgish),
\textsc{lim}~(Limburgish),
\textsc{ksh}~(Colognian),
\textsc{pfl}~(Palatine German),
\textsc{pdc}~(Pennsylvania Dutch),
\textsc{yid}~(Yiddish), 
\textsc{sxu}~(Upper Saxon),
\textsc{vmf}~(East Franconian),
\textsc{bar}~(Bavarian),
\textsc{swg}~(Swabian),
\textsc{gsw}~(Swiss German and Alsatian),
\textsc{wae}~(Walser),
and
\textsc{cim}~(Cimbrian).
Our survey also encompasses data for dialects/non-standard varieties of
Norwegian~(\textsc{nor}),
Swedish~(\textsc{swe}),
Danish~(\textsc{dan}),
English~(\textsc{eng}),
and German~(\textsc{deu})
that do not have their own ISO codes.

We use ISO codes to refer to (groups of) language varieties for practical reasons -- despite their shortcomings as labels for varieties from linguistic continua \cite{morey2013iso, nordhoff2011glottolog}, they are widely used and recognized, and many of the corpora in this survey are described in terms that easily map to ISO codes.

In some cases, the codes or the corpus descriptions are ambiguous. 
For instance, many Low Saxon corpora contain entries that also belong to one of the more specific Dutch Low Saxon codes, and some Swiss German corpora also contain some Walser content.
Where possible (and where the data instances themselves are labelled on a precise enough level), we use the more specific codes.

Table~\ref{tab:by-language} provides an overview of resource types by language variety.

\begin{table*}
\begin{adjustbox}{max width=\textwidth, center}
\begin{tabular}{ll@{}llllll@{\minicolspace}l@{\minicolspace}l@{\minicolspace}l@{\minicolspace}l@{\minicolspace}l}
\toprule
\multicolumn{2}{l}{\textbf{Language}} & 
\begin{tabular}[c]{@{}l@{}}\textbf{Dialect/}\\\textbf{Location}\end{tabular} & 
\textbf{\begin{tabular}[c]{@{}l@{}}Morpho-\\ syntax\end{tabular}} & 
\textbf{Semantic} & 
\textbf{\begin{tabular}[c]{@{}l@{}}Parallel\\(curated)\end{tabular}} &  
\textbf{\begin{tabular}[c]{@{}l@{}}Uncurated\\ text\end{tabular}} & 
\multicolumn{5}{l}{\textbf{\begin{tabular}[c]{@{}l@{}}Curated\\data\end{tabular}}} \\
\multicolumn{2}{l}{\textit{North Germanic}}\minivskip\\
\textsc{fao} & Faroese & \pinMorphoSemParUncur{\pin}{\yes}{\no}{\no}{\yes} \formats{\audio}{\phonetic}{}{\ownOrtho}{} \\
\textsc{nor} & (non-std.) Norwegian & \pinMorphoSemParUncur{\pin}{\yes}{\no}{\no}{\no} \formats{\audio}{\phonetic}{\doculectOrtho{}}{}{\normalized{}} \minivskip\\
\textsc{ovd} & Elfdalian & \pinMorphoSemParUncur{\pin}{\no}{\no}{\no}{\no} \formats{}{}{}{\ownOrtho}{\normalized{}}\\
\textsc{swe} & (non-std.) Swedish & \pinMorphoSemParUncur{\pin}{\no}{\no}{\no}{\no} \formats{}{\phonetic}{}{}{\normalized{}} \\
\textsc{dan} & (non-std.) Danish & \pinMorphoSemParUncur{\pin}{}{}{}{\yes} \formats{}{}{\textbf{?}}{}{} \minivskip\\

\multicolumn{2}{l}{\textit{Anglo-Frisian}}\minivskip\\
\textsc{sco} & Scots & \pinMorphoSemParUncur{}{\yes}{}{}{\yes} \formats{}{}{\doculectOrtho{}}{\ownOrtho}{\normalized{}}\\\minivskip
\textsc{eng} & (non-std.) English & \pinMorphoSemParUncur{\pin}{\yes}{}{}{} \formats{\audio}{}{}{}{\normalized{}} \\
\textsc{fry} & West Frisian & \pinMorphoSemParUncur{\pin}{\yes}{}{}{\yes} \formats{\audio}{}{}{\ownOrtho}{}\\
\textsc{frr} & North Frisian & \pinMorphoSemParUncur{}{}{}{}{\yes} \formats{}{}{}{}{}\\\minivskip
\textsc{stq} & Saterland Frisian & \pinMorphoSemParUncur{}{}{}{}{\yes} \formats{}{}{}{}{}\\

\multicolumn{2}{l}{\textit{Low German*}}\minivskip\\
\textsc{nds} & Low Saxon & \pinMorphoSemParUncur{\pin}{\yes}{\no}{\yes}{\yes} \formats{\audio}{}{\doculectOrtho{}}{\ownOrtho}{}\\
\textsc{frs} & East Frisian Low Saxon &\pinMorphoSemParUncur{}{}{}{}{\yes} \formats{\audio}{}{}{}{}\\
\textsc{gos} & Gronings & \pinMorphoSemParUncur{}{}{}{\yes}{\yes} \formats{}{}{}{}{}\minivskip\\
\textsc{twd} & Twents & \pinMorphoSemParUncur{}{}{}{}{\yes} \formats{}{}{\doculectOrtho}{}{}
\\
\textsc{act} & Achterhoeks & \pinMorphoSemParUncur{}{}{}{}{\yes} \formats{}{}{\doculectOrtho}{}{}
\\
\textsc{wep} & Westphalian & \pinMorphoSemParUncur{}{}{}{}{} \formats{\audio}{}{}{}{\normalized}\minivskip
\\

\multicolumn{2}{l}{\textit{Macro-Dutch}}\minivskip\\
\textsc{vls} & West Flemish &  \pinMorphoSemParUncur{\pin}{\yes}{}{}{\yes} \formats{}{\phonetic}{}{}{}\\\minivskip
\textsc{zea} & Zeelandic & \pinMorphoSemParUncur{}{}{}{}{\yes} \formats{}{}{}{}{} \\

\multicolumn{2}{l}{\textit{Middle German}}\minivskip\\
\textsc{ltz} & Luxembourgish & \pinMorphoSemParUncur{}{}{}{}{\yes} \formats{}{}{}{\ownOrtho}{}\\
\textsc{ksh} & Colognian & \pinMorphoSemParUncur{}{}{}{}{\yes} \formats{}{}{}{}{}\\\minivskip
\textsc{lim} & Limburgish & \pinMorphoSemParUncur{}{}{}{}{\yes} \formats{}{}{}{}{} \\
\textsc{pfl} & Palatine German & \pinMorphoSemParUncur{}{}{}{}{\yes} \formats{}{}{}{}{}\\
\textsc{pdc} & Pennsylvania Dutch & \pinMorphoSemParUncur{}{}{}{}{\yes} \formats{}{}{}{}{}\\
\textsc{yid} & Yiddish** & \pinMorphoSemParUncur{}{\yes}{}{}{\yes} \formats{\audio}{}{\phonetic}{}{}\\
\textsc{sxu} & Upper Saxon & \pinMorphoSemParUncur{}{}{}{}{} \formats{\audio}{}{}{}{\normalized}\minivskip\\

\multicolumn{2}{l}{\textit{Upper German}}\minivskip\\
\textsc{deu} & (non-std.) German & \pinMorphoSemParUncur{}{}{}{\yes}{} \formats{\audio}{}{\phonetic}{}{\normalized} \\
\textsc{vmf} & East Franconian & \pinMorphoSemParUncur{}{}{}{}{} \formats{\audio}{}{}{}{\normalized}\\
\textsc{bar} & Bavarian & \pinMorphoSemParUncur{}{\yes}{\yes}{\yes}{\yes} \formats{\audio}{\phonetic}{\doculectOrtho}{}{\normalized}\\\minivskip
\textsc{cim} & Cimbrian & \pinMorphoSemParUncur{}{}{}{}{} \formats{\audio}{}{}{}{\normalized}\\
\textsc{swg} & Swabian & \pinMorphoSemParUncur{}{}{}{}{\yes} \formats{}{}{}{}{}\\
\textsc{gsw} & Swiss Ger. \& Alsatian & \pinMorphoSemParUncur{\pin}{\yes}{\yes}{\yes}{\yes} \formats{\audio}{\phonetic}{\doculectOrtho}{}{\normalized}\\
\textsc{wae} & Walser & \pinMorphoSemParUncur{\pin}{}{\yes}{\yes}{\yes} \formats{\audio}{}{\doculectOrtho}{}{} \\
\bottomrule
\end{tabular}

\end{adjustbox}
\caption{\textbf{Corpora by language variety.} For ease of reference, the language are sorted by Germanic subbranches (based on Glottolog \cite{glottolog-4-7}). 
*For additional texts written in varieties of Low German/Saxon with other ISO 639-3 codes, see the note on the Low Saxon Wikipedias in Table~\ref{tab:wikistats}.
**Glottolog discerns between Eastern Yiddish (Middle German) and Western Yiddish (Upper German).
Symbols:
\audio{}~=~audio,
\phonetic{}~=~phonetic/phonemic transcription,
\doculectOrtho{}~=~pronunciation spelling, 
\ownOrtho{}~=~LRL orthography,
\normalized{}~=~normalized orthography.
}
\label{tab:by-language}
\end{table*}

\section{Written representations}
\label{sec:transcriptions}

\begin{table*}
\begin{adjustbox}{max width=\textwidth, center}
\begin{tabular}{lll}
\toprule
\multicolumn{3}{l}{From the Faroese BLARK recordings \cite{simonsen2022faroese-blark}:}\\
1a & \ownOrtho & vit høvdu matpakka við og eg hugnaði mær óført\\
1b & \phonetic & vId h9dI m\%EApaHga v\%i: o e h\%u:najI mar \%OW:f9zd \\
1c & \phonetic & v\textsci{}d h\oe{}d\textsci{} \textprimstress{}m\textepsilon{}\textsubarch{a}p\textsuperscript{h}a\textsuperscript{h}ga \textprimstress{}vi\textlengthmark{} o e \textprimstress{}hu\textlengthmark{}naj\textsci{} ma\textturnr{} \textprimstress{}\textopeno{}\textsubarch{u}\textlengthmark{}f\oe{}\textrtails{}d \\
\multicolumn{3}{l}{``We had lunchboxes with us and I enjoyed myself greatly.''}\\\midrule

\multicolumn{3}{l}{From the Norwegian NB Tale corpus \cite{nb-tale}:}\\
2a & \normalized & Etter litt godsnakk kom tre av kyrne mot han mens den fjerde glei og fall \\
2b & \phonetic & ""\{t@4 l"it g""u:snAkk k"Om t4"e: "A:v C"y:n`{}@ m"u:t "An m"ens d\_= fj""\{:d`{}@ gl"eI "O: f"Al\\
2c & \phonetic & \textsuperscript{2}\textepsilon{}t\textschwa{}\textturnr{} \textsuperscript{1}l\textsci{}t \textsuperscript{2}gu\textlengthmark{}sn\textscripta{}kk \textsuperscript{1}k\textopeno{}m \textsuperscript{1}t\textturnr{}e\textlengthmark{} \textsuperscript{1}\textscripta{}\textlengthmark{}v \textsuperscript{1}\c{c}y\textlengthmark{}\textrtailn{}\textschwa{} \textsuperscript{1}mu\textlengthmark{}t \textsuperscript{1}\textscripta{}n \textsuperscript{1}m\textepsilon{}ns d\textsyllabic{n} \textsuperscript{2}fj\ae{}\textlengthmark{}\textrtaild{}\textschwa{} \textsuperscript{1}gl\textepsilon{}\textsubarch{\textsci{}} \textsuperscript{1}o\textlengthmark{} \textsuperscript{1}f\textscripta{}l\\
\multicolumn{3}{l}{``After some coaxing, three of the cows came towards him while the fourth one slipped and fell.''}\\
\midrule

\multicolumn{3}{l}{From the Norwegian part of the Nordic Dialect Corpus \cite{johannessen2009nordic-dialect-corpus}:}\\
3a & \normalized & det er slik at de fleste kommer jo att når de får \# unger \\
3b & \phonetic & de e sjlik att dæi fLeste kjemme jo att nårr dæi fær \# onnga  \\
\multicolumn{3}{l}{``The thing is that most people return when they have \textit{[brief pause]} kids.''}\\\midrule

\multicolumn{3}{l}{From the Elfdalian part of the Nordic Dialect Corpus \cite{johannessen2009nordic-dialect-corpus}:}\\
4a & \ownOrtho & wen wa wen war e\dh{} f\o{}r ien måna\dh{} ? juni ? \\
4b & \normalized & vad va- vad var det för en månad ? juni ? \\
\multicolumn{3}{l}{``What, wha-, what month was it? June?''}\\
\midrule

\multicolumn{3}{l}{From UD Low Saxon LSDC \cite{siewert2021lsdc-with-ud}:}\\
5a & \ownOrtho & Nu leyt em de böyse vynd disse nacht kyn ouge an enander doon. \\
5b & \doculectOrtho & Nu leit em de baise Find düse Nacht kinn Auge an enander dohn. \\
\multicolumn{3}{l}{``Now the wicked enemy didn't let them get a wink of sleep that night.''}\\\midrule

\multicolumn{3}{l}{From the Swiss German ArchiMob corpus \cite{Scherrer2019ArchiMob}:}\\
6a & \normalized & können sie ihre jugendzeit beschreiben \\
6b & \phonetic & chönd sii iri jugendziit beschriibe \\
\multicolumn{3}{l}{``Can you describe your youth?''}\\
\midrule

\multicolumn{3}{l}{From the BISAME corpus \cite{bisame}:}\\
7a & \doculectOrtho & Niema hat salamols gweßt as die Werter vum franzeescha kumma.\\
\multicolumn{3}{l}{``Nobody knew then that these words came from French.''}\\

\bottomrule
\end{tabular}

\end{adjustbox}
\caption{%
\textbf{Examples of written representations.}
Symbols:
\phonetic{}~=~phonetic/phonemic transcription,
\doculectOrtho{}~=~pronunciation spelling, 
\ownOrtho{}~=~LRL orthography,
\normalized{}~=~normalized orthography.
}
\label{tab:transcriptions}
\end{table*}

Table~\ref{tab:transcriptions} provides examples of different types of written representations and showcases how diverse each category can be.

Examples 1a, 2a, 3a, 4a/b, 5a and 6a are written in \textbf{standardized orthographies} (or in lower-cased versions of standard orthographies with no pronunciation).
Of these, sentences 1a, 4a and 5a are written in orthographies developed for their respective low-resource languages~\ownOrtho, while 2a, 3a, 4b and 6a are normalized and written in the orthographies of closely related standard languages~\normalized{} (the last two are Elfdalian written in Swedish and Swiss German written in Standard German).

Sentences 5b and 7a present two examples of ad-hoc \textbf{pronunciation spellings}~\doculectOrtho.
These kinds of spellings vary from speaker to speaker, and one and the same speaker might also choose different spellings of the same word at different times.

\textbf{Phonetic or phonemic transcriptions}~\phonetic{} have different styles depending on each corpus's transcription guidelines.
Examples 1b and 2b are written in modified versions of SAMPA and X-SAMPA, and the corpora come with sufficient documentation to automatically convert these transcriptions into IPA (1c, 2c).
(The superscript symbols~\textsuperscript{1} and~\textsuperscript{2} in example 2c are commonly used to indicate the Norwegian pitch accent.)
The transcription styles presented in examples 3b and 6b are based on Norwegian and Standard German orthography, respectively.
What sets them apart from pronunciation spellings is that they are consistent across the entire corpus and that they follow linguistic rationales that often are outlined in the corpus documentation.

\section{Overlapping data sources}
\label{sec:overlaps}

Several of the corpora mentioned in this article overlap with each other:

\begin{itemize}
    \item \textit{UD Faroese OFT} and 
    the \textit{Korp} subcorpus of the background corpus of the Faroese \textit{BLARK 1.0} contain material from the Faroese Wikipedia.
    \item The \textit{NDC Treebank} uses data from the \textit{Nordic Dialect Corpus.}
    \item The \textit{LIA Treebank} and \textit{UD Norwegian NynorskLIA} are annotated subsets of \textit{LIA Norsk}, and they overlap with each other.
    \item The \textit{POS-tagged Scots corpus} contains annotated sentences from \textit{SCOTS.}
    \item \textit{UD Low Saxon LSDC} and \textit{LSDC} overlap.
    \item \textit{UD Frisian/Dutch Fame} is an annotated subset if \textit{FAME}.
    \item Many of the sentences in \textit{UD Swiss German UZH} are also in \textit{NOAH's corpus.} Both of these corpora contain material from the Alemannic Wikipedia.
    \item \textit{SB-CH} contains \textit{NOAH's corpus.}
    \item The \textit{Annotated Corpus for the Alsatian Dialects} contains articles from the Alemannic Wikipedia that were explicitly tagged as Alsatian.
    \item \textit{TaPaCo} is a subset of \textit{Tatoeba.}
    \item Any corpus that includes data from the internet might overlap with the uncurated datasets in Section~\ref{sec:uncurated}.
\end{itemize}

\section{Wikipedia statistics}
\label{sec:wikistats}

Table~\ref{tab:wikistats} provides a comparison of Wikipedia sizes and user (vs.\ bot) activity.\footnote{%
The data sources are the automatically updated list of Wikipedia sizes at \href{https://meta.wikimedia.org/wiki/List_of_Wikipedias_by_language_group\#Germanic}{\texttt{meta.wikimedia.org/\allowbreak{}wiki/\allowbreak{}List\_\allowbreak{}of\_\allowbreak{}Wikipedias\_\allowbreak{}by\_\allowbreak{}language\_\allowbreak{}group\allowbreak{}\#Germanic}} (last accessed 2023-01-31) and Wikimedia's metadata via \https{wikimedia.org/api/rest\_v1/}.
The scripts are available via \href{https://github.com/mainlp/wikistats}{\texttt{github.com/\allowbreak{}mainlp/\allowbreak{}wikistats}}.}
The sizes of the small Germanic Wikipedias vary considerably from wiki to wiki (there are just under 2k Pennsylvania Dutch articles, while the (German) Low Saxon Wikipedia has over 84k articles), as does the number of recently active contributors (from 6 active non-bot users per month for Ripurarian/Colognian, Palatine German and Pennsylvania Dutch to 70 for Scots).

While bots can be used for automating many tasks that are unrelated to the textual diversity of a wiki (e.g., cleaning up article redirection pages), they can also be used to automatically create short template-based articles.\footnote{For an example for the latter, see \href{https://nds.wikipedia.org/wiki/Bruker:ArtikelBot}{\texttt{nds.wikipedia.org/\allowbreak{}wiki/\allowbreak{}Bruker:ArtikelBot}}}
The share of manual edits (i.e., edits not by bots) is very varied across wikis -- only about a quarter of all edits in the Pennsylvania Dutch Wikipedia have been made manually, compared to 79~\% in the North Frisian Wikipedia.
However, there is a clear trend towards a much larger proportion of manual edits: the vast majority of edits made only in the past year were manual edits.

Some of the wikis are written according to one or more orthographies,
while others either do not include any spelling recommendations at all or encourage editors to use whatever pronunciation spelling they prefer.
The Dutch Low Saxon Wikipedia, for instance, recommends \textit{Nysassiske Skryvwyse}, whereas the German Low Saxon Wikipedia recommends another orthography: \textit{Sass-Schrievwies}.
The Ripurarian/Colognian Wikipedia, conversely, encourages idiosyncratic spellings.\footnote{%
These are the pages detailing orthographic conventions we were able to find (sorted by wiki size):
\href{https://nds.wikipedia.org/wiki/Wikipedia:Sass}{\texttt{nds.wikipedia.org/wiki/Wikipedia:Sass}};
\href{https://sco.wikipedia.org/wiki/Wikipedia:Spellin_an_grammar}{\texttt{sco.wikipedia.org/\allowbreak{}wiki/\allowbreak{}Wikipedia:\allowbreak{}Spellin\_an\_grammar}};
\href{https://als.wikipedia.org/wiki/Hilfe:Schrybig}{\texttt{als.wikipedia.org/\allowbreak{}wiki/\allowbreak{}Hilfe:\allowbreak{}Schrybig}};
\href{https://bar.wikipedia.org/wiki/Wikipedia:Wia\_schreib\_i\_a\_guads\_Boarisch\%3F}{\texttt{bar.wikipedia.org/\allowbreak{}wiki/\allowbreak{}Wikipedia:\allowbreak{}Wia\_\allowbreak{}schreib\_\allowbreak{}i\_\allowbreak{}a\_\allowbreak{}guads\_\allowbreak{}Boarisch\%3F}};
\href{https://frr.wikipedia.org/wiki/Wikipedia:Spr\%C3\%A4kekoordinasjoon}{\texttt{frr.wikipedia.org/\allowbreak{}wiki/\allowbreak{}Wikipedia:\allowbreak{}Spräkekoordinasjoon}};
\href{https://li.wikipedia.org/wiki/Wikipedia:Wie\_sjrief\_ich\_Limburgs}{\texttt{li.wikipedia.org/\allowbreak{}wiki/\allowbreak{}Wikipedia:\allowbreak{}Wie\_\allowbreak{}sjrief\_\allowbreak{}ich\_\allowbreak{}Limburgs}};
\href{https://vls.wikipedia.org/wiki/Wikipedia:Gebruuk\_van\_streektoaln}{\texttt{vls.wikipedia.org/\allowbreak{}wiki/\allowbreak{}Wikipedia:\allowbreak{}Gebruuk\_\allowbreak{}van\_\allowbreak{}streektoaln}};
\href{https://nds-nl.wikipedia.org/wiki/Wikipedia:Spelling}{\texttt{nds-nl.\allowbreak{}wikipedia.org/\allowbreak{}wiki/\allowbreak{}Wikipedia:\allowbreak{}Spelling}};
\href{https://stq.wikipedia.org/wiki/Wikipedia:Hälpe\_bie\_ju\_seelter\_Sproake}{\texttt{stq.wikipedia.org/\allowbreak{}wiki/}}
\href{https://stq.wikipedia.org/wiki/Wikipedia:Hälpe\_bie\_ju\_seelter\_Sproake}{\texttt{Wikipedia:\allowbreak{}Hälpe\_\allowbreak{}bie\_\allowbreak{}ju\_\allowbreak{}seelter\_\allowbreak{}Sproake}};
\href{https://ksh.wikipedia.org/wiki/Wikipedia:Schrievwies}{\texttt{ksh.wikipedia.org/\allowbreak{}wiki/\allowbreak{}Wikipedia:\allowbreak{}Schrievwies}}
}

Several of these wikis include (some) articles with metadata specifying which variety the document is written in.\footnote{%
\label{note-tags}\raggedright Sorted by wiki size:
\href{https://nds.wikipedia.org/wiki/Kategorie:Artikels\_na\_Dialekt}{\tt nds.wikipedia.org/wiki/\allowbreak{}Kategorie:\allowbreak{}Artikels\_\allowbreak{}na\_Dialekt}; 
\href{https://als.wikipedia.org/wiki/Kategorie:Wikipedia:Dialekt}{\tt als.\allowbreak{}wikipedia.\allowbreak{}org/\allowbreak{}wiki/\allowbreak{}Kategorie:\allowbreak{}Wikipedia:\allowbreak{}Dialekt};
\href{https://bar.wikipedia.org/wiki/Kategorie:Artikel\_nach\_Dialekt}{\tt bar.\allowbreak{}wikipedia.\allowbreak{}org/\allowbreak{}wiki/\allowbreak{}Kategorie:\allowbreak{}Artikel\_\allowbreak{}nach\_\allowbreak{}Dialekt};
\href{https://frr.wikipedia.org/wiki/Kategorie:Spriakwiisen}{\tt frr.\allowbreak{}wikipedia.\allowbreak{}org/\allowbreak{}wiki/\allowbreak{}Kategorie:\allowbreak{}Spriakwiisen};
\href{https://li.wikipedia.org/wiki/Categorie:Wikipedia:Artikele\_nao\_dialek}{\tt li.\allowbreak{}wikipedia.\allowbreak{}org/\allowbreak{}wiki/\allowbreak{}Categorie:\allowbreak{}Wikipedia:\allowbreak{}Artikele\_\allowbreak{}nao\_\allowbreak{}dialek};
\href{https://vls.wikipedia.org/wiki/Categorie:Wikipedia:Artikels\_noar\_dialect}{\tt vls.\allowbreak{}wikipedia.\allowbreak{}org/\allowbreak{}wiki/\allowbreak{}Categorie:\allowbreak{}Wikipedia:\allowbreak{}Artikels\_\allowbreak{}noar\_\allowbreak{}dialect};
\href{https://nds-nl.wikipedia.org/wiki/Kategorie:Nedersaksies\_artikel}{\tt nds-nl.\allowbreak{}wikipedia.\allowbreak{}org/\allowbreak{}wiki/\allowbreak{}Kategorie:\allowbreak{}Nedersaksies\_\allowbreak{}artikel};
\href{https://ksh.wikipedia.org/wiki/Saachjrupp:Wikipedia:Atikkel\_ier\_Shprooche}{\tt ksh.\allowbreak{}wikipedia.\allowbreak{}org/\allowbreak{}wiki/\allowbreak{}Saachjrupp:\allowbreak{}Wikipedia:\allowbreak{}Atikkel\_\allowbreak{}ier\_\allowbreak{}Shprooche};
\href{https://pfl.wikipedia.org/wiki/Sachgrubb:Adiggel\_noch\_em\_Dialegd}{\tt pfl.\allowbreak{}wikipedia.\allowbreak{}org/\allowbreak{}wiki/\allowbreak{}Sachgrubb:\allowbreak{}Adiggel\_\allowbreak{}noch\_\allowbreak{}em\_\allowbreak{}Dialegd}
}

\begin{table*}[t]
\begin{adjustbox}{max width=\textwidth, center}

\begin{tabular}{ll@{}S[table-format=4.0]@{\minicolspace}lS[table-format=2.0]@{\minicolspace}lS[table-format=2.0]@{\minicolspace}lS[table-format=6.0]}
\toprule
\multicolumn{2}{l}{\textbf{Wikipedia \& Language}} & \multicolumn{2}{l}{\textbf{\begin{tabular}[c]{@{}l@{}}Articles\\ (01/2023)\end{tabular}}} & \multicolumn{2}{l}{\textbf{\begin{tabular}[c]{@{}l@{}}Manual\\ edits\\ (2001--2022)\end{tabular}}} & \multicolumn{2}{l}{\textbf{\begin{tabular}[c]{@{}l@{}}Manual\\ edits\\ (2022)\end{tabular}}} & {\textbf{\begin{tabular}[c]{@{}r@{}}Monthly\\ editors\\ (2022)\end{tabular}}}  \\
\midrule
nds & \textsc{nds} (Germany)* (\pin) & 84 & k & 44 & \% & 99 & \% & 30 \\
lb & \textsc{ltz} & 61 & k & 43 & \% & 85 & \% & 56 \\
fy & \textsc{fry} & 50 & k & 60 & \% & 99 & \%  & 54 \\\minivskip
sco & \textsc{sco} & 39 & k & 53 & \%  & 63 & \%  & 70 \\
als & \textsc{gsw + swg + wae} (\pin) & 30 & k & 69 & \%  & 100 & \%  & 58 \\
bar & \textsc{bar} (\pin) & 27 & k & 68 & \%  & 63 & \%  & 39 \\
frr & \textsc{frr} (\pin) & 17 & k & 79 & \%  & 85 & \%  & 16 \\\minivskip
yi & \textsc{yid} & 15 & k & 49 & \% & 97 & \% & 35 \\
li & \textsc{lim} & 14 & k & 42 & \%  & 75 & \%  &  21 \\
fo & \textsc{fao} & 14 & k & 41 & \% & 99 & \% & 29 \\
vls & \textsc{vls} (\pin) & 8 & k & 45 & \% & 79 & \% & 16 \\\minivskip
nds-nl & \textsc{nds} (Netherlands)* (\pin) & 8 & k & 40 & \%  & 68 & \%  & 14\\
zea & \textsc{zea} & 6 & k & 47 & \%  & 98 & \%  & 10 \\
stq & \textsc{stq} & 4 & k & 38 & \%  & 81 & \%  & 8 \\
ksh &  \textsc{ksh} + other Ripuarian (\pin) & 3 & k & 32 & \%  & 99 & \%  & 6 \\
pfl & \textsc{pfl} + oth. Rhen. Franc., Hessian (\pin) & 3 & k & 65 & \% & 72 & \% & 6 \\
pdc & \textsc{pdc} & 2 & k & 27 & \% & 92 & \%  & 6 \\
 \midrule
en & \textsc{eng} & 6608 & k & 90 & \% & 92 & \% & 102574 \\
de & \textsc{deu} & 2765 & k & 91 & \% & 93 & \% & 16141 \\
nl & \textsc{nld} & 2114 & k & 68 & \%  & 66 & \% & 3521 \\
da & \textsc{dan} & 289 & k & 63 & \%  & 64 & \% &  711 \\
is & \textsc{isl} & 56 & k & 54 & \%  & 79 & \% & 118  \\
\bottomrule
\end{tabular}

\end{adjustbox}
\caption{\textbf{Wikipedia statistics.}
`Manual edits' include the proportion of edits (of content pages) performed by registered non-bot users or anonymous editors (out of the total number of content page edits performed by anyone, including bots).
The number of monthly editors is the mean number of registered non-bot users who edited at least one content page, per month.
English, German, Dutch (NLD), Danish (DAN) and Icelandic (ISL) are included for comparison.
The wikis with a pin symbol~\pin{} contain (some) articles tagged by dialect; see footnote~\ref{note-tags}.
*The \textit{nds} and \textit{nds-nl} wikis are primarily concerned with varieties of Low Saxon spoken in, respectively, Germany and the Netherlands.
The former also contains articles written in varieties associated with the ISO 639-3 codes \textsc{wep} and \textsc{frs}, 
and the latter with \textsc{act}, \textsc{frs}, \textsc{gos}, \textsc{drt} (Drents), \textsc{sdz} (Sallands), \textsc{stl} (Stellingwerfs), \textsc{twd} and \textsc{vel} (Veluws).
}
\label{tab:wikistats}
\end{table*}

\end{document}